%% file: acl_latex.tex
\documentclass[11pt]{article}

\usepackage[final]{acl}

\usepackage{times}
\usepackage{latexsym}

\usepackage[T1]{fontenc}

\usepackage[utf8]{inputenc}

\usepackage{microtype}

\usepackage{inconsolata}

\usepackage{graphicx}
\usepackage{amsmath}
\usepackage{booktabs}
\usepackage{multirow}
\usepackage{enumitem}
\usepackage{amsfonts}
\title{LoopMoE: Unifying Iterative Computation with Mixture-of-Experts for Language Modeling}

\author{
 \textbf{Wenkai Chen\textsuperscript{1}},
 \textbf{Tianshu Li\textsuperscript{2}},
 \textbf{Wenyong Huang\textsuperscript{2}},
 \textbf{Yichun Yin\textsuperscript{2}},
\\
 \textbf{Lifeng Shang\textsuperscript{2}},
 \textbf{Chengwei Qin\textsuperscript{1}}
\\
\\
 \textsuperscript{1}Hong Kong University of Science and Technology (Guangzhou)\\
 \textsuperscript{2}Huawei Technologies Co., Ltd.
\\
 \small{
   \textbf{Correspondence:} \href{mailto:email@domain}{chengweiqin@hkust-gz.edu.cn}
 }
}
\newcommand{\ourmodel}{LoopMoE }
\newcommand{\ourmodelnospace}{LoopMoE}

\begin{document}
\maketitle
\begin{abstract}
\input{sections/0-abstract.tex}

\end{abstract}

\section{Introduction}
\label{sec:intro}
\input{sections/1-introduction}

\section{Related Work}
\label{sec:bg}
\input{sections/2-related_work}

\section{Methods}
\label{sec:method}
\input{sections/3-method}

\section{Experimental Setup}
\label{sec:exp_setup}
\input{sections/4-experimental_setup}

\section{Results}
\label{sec:results}
\input{sections/5-results}

\section{Analysis}
\label{sec:analysis}
\input{sections/6-analysis}

\section{Conclusion}
\label{sec:conclusion}
\input{sections/7-conclusion}

\section*{Limitations}
\label{sec:limitations}
\input{sections/limitations}

\bibliography{custom}

\appendix

\section{Appendix}
\label{sec:appendix}
\input{sections/appendix}

\end{document}

%% file: sections/0-abstract.tex
Mixture-of-Experts (MoE) and looped architectures scale models along two orthogonal axes, namely parameter capacity and effective depth. However, mainstream looped architectures rely on dense backbones that couple parameter count with per-token FLOPs, which makes it impossible to isolate the effect of iterative computation under matched budgets. To this end, we present LoopMoE, a looped MoE language model that integrates sparse routing with iterative weight-shared computation through two designs. The first is IterAdaLN, which resolves weight-sharing symmetry via a modulation signal jointly conditioned on the iteration index and the per-token hidden state. The second is a capacity-balancing strategy that recovers the attention-to-FFN active parameter ratio of well-tuned non-looped references. Together, these designs enable the first strictly controlled, head-to-head evaluation of a looped MoE against a Vanilla MoE under identical total parameters, per-token FLOPs, and active sublayer ratios. Across nine downstream benchmarks, LoopMoE’s average improvement over its matched vanilla MoE increases from over 1 point at the 3B scale to approximately 3 points at the 9B scale. These results provide initial evidence that the benefits of iterative sparse computation may strengthen with scale, positioning LoopMoE as a promising architecture for scalable looped language models.

%% file: sections/1-introduction.tex
Capability scaling in modern large language models (LLMs) has been predominantly driven by parameter expansion, with Mixture-of-Experts (MoE) architectures emerging as the de facto standard for achieving this at a tractable cost~\cite{liu2024deepseek, yang2025qwen3, zeng2025glm}. The efficacy of MoE stems from a structural decoupling, in which activating only a sparse subset of experts per token disaggregates total parameter count from per-token active compute and enables parameter growth without commensurate FLOP inflation. However, scaling total parameters is not the sole avenue for enhancing capability per unit of training cost. An orthogonal paradigm investigates how to extract greater computational depth from a strictly bounded parameter budget. Among these approaches, block-recurrent or looped architectures offer a compelling solution by executing a shared block of layers over multiple iterations, thereby increasing depth without introducing new parameters~\cite{dehghani2018universal, lan2019albert, bae2025relaxed, mohtashami2025cotformer, bae2026mixture, geiping2026scaling}.

While block-recurrent architectures successfully increase computational depth, existing dense loops~\cite{zhu2025scaling, zeitoun2026hyperloop, jeddi2026loopformer} suffer from a fundamental structural flaw: parameter count and per-token FLOPs remain tightly coupled. Iterating a dense block multiplies compute without altering parameters, which makes it impossible to simultaneously match a non-looped baseline on both axes. MoE architectures~\cite{cai2025survey}, which naturally decouple total parameters from active compute, offer a theoretical solution. However, naively integrating loops into an MoE backbone introduces two critical new challenges. Representationally, weight sharing across iterations imposes structural symmetry that severely restricts the model's ability to evolve token states progressively. Structurally, it induces asymmetrical active parameter expansion: attention parameters are reused uniformly across iterations, whereas tokens dynamically route to different experts at each step. As a consequence, the active attention-to-FFN parameter ratio $\rho$ deviates substantially from the operating point $\rho^\star$ of well-tuned baselines.

To fully unlock the potential of iterative sparse computation, we introduce \textbf{\ourmodelnospace}, a novel block-recurrent MoE language model designed to overcome both limitations. \ourmodel adopts a streamlined sandwich layout in which the core loop body evolves through pure recursion. To break the structural symmetry inherent in weight sharing, we introduce \textbf{IterAdaLN}, a token-level modulation scheme inspired by adaptive normalization in conditional generation~\cite{perez2018film, peebles2023scalable}. Operating strictly at a token-level granularity, IterAdaLN dynamically generates affine parameters jointly from the iteration index and the per-token hidden state. Replacing static RMSNorm~\cite{zhang2019root} with this token-level conditioning ensures sufficient differentiation across iterations. It also allows the representational trajectory to remain highly expressive without requiring iteration-specific prefix re-injection.

Crucially, \ourmodel leverages the inherent flexibility of the MoE framework to resolve the asymmetrical expansion problem. To maintain the optimal ratio $\rho^\star$, we adopt a principled balancing strategy that expands the Q/KV LoRA ranks of Multi-head Latent Attention (MLA)~\cite{liu2024deepseek} and correspondingly reduces the expert hidden dimensions. This decoupling allows independent adjustment of attention and FFN capacities. We then recover the baseline's total parameter count $N^\star$ by scaling the routed expert count. Together, these designs turn the MoE structure into the exact degree of freedom needed to construct a rigorously controlled looped architecture.

Our main contributions are summarized below:
\begin{itemize}[leftmargin=*]
\item \textbf{Novel Architecture}: We propose \textbf{\ourmodelnospace}, which couples sparse expert routing with iterative weight-shared computation through two key designs: \textbf{IterAdaLN}, a token-level iteration-conditioned modulation that breaks weight-sharing symmetry, and a capacity-balancing strategy that maintains the attention-to-FFN active-parameter ratio of well-tuned non-looped MoEs.
\item \textbf{Strictly Controlled Evaluation}: We provide the first head-to-head comparison of a looped architecture against a Vanilla MoE under rigorously matched total parameters, per-token FLOPs, and active sublayer ratios, cleanly isolating the architectural contribution from parameter or compute inflation.
\item \textbf{Consistent Gains with Positive Scaling Behavior}: Under strictly matched budgets, \ourmodelnospace{} outperforms the vanilla MoE on 8 of 9 benchmarks at the 3B scale, with an average improvement exceeding 1 point despite using fewer unique physical parameters per forward pass. At the 9B scale, its average improvement over the matched vanilla MoE increases to nearly 3 points. This widening gap suggests positive scaling behavior, whereby the benefits of iterative sparse computation become more pronounced as model capacity grows.
\end{itemize}

%% file: sections/2-related_work.tex
\subsection{Mixture-of-Experts Language Models}
Mixture-of-Experts has become a standard approach for scaling LLMs under a controlled compute budget~\cite{lepikhin2020gshard, fedus2022switch, cai2025survey}. Recent systems explore fine-grained experts, shared experts, and auxiliary-loss-free load balancing~\cite{jiang2024mixtral, xue2024openmoe, liu2024deepseek, yang2025qwen3, zeng2025glm}. These architectures all assume standard non-shared per-layer parameters, and their interaction with iterative weight sharing remains largely unexplored. Our work targets exactly this intersection.

\subsection{Looped and Weight-Shared Transformers}

Looped and weight-shared architectures decouple parameter count from effective depth by reusing the same parameters for iterative refinement~\cite{hutchins2022block}. Universal Transformer~\cite{dehghani2018universal} ties a single block across depth and applies per-token adaptive halting, whereas ALBERT~\cite{lan2019albert} shares one block across all layers of an otherwise standard encoder purely for parameter efficiency. More recent designs revisit input-dependent depth and halting~\cite{bae2025relaxed,bae2026mixture}, explicit iterative reasoning~\cite{mohtashami2025cotformer}, and conditional modulation of shared weights across iterations~\cite{jeddi2026loopformer}. 
A closely related line of work also studies latent reasoning through recurrence in the hidden state~\cite{hao2024training, saunshi2025reasoning}.
A separate line of work~\cite{mcleish2025teaching, zeitoun2026hyperloop, geiping2026scaling} wraps a recurrent middle block with non-looped layers on either side, a prefix-loop-suffix design that we also adopt.

\subsection{Adaptive LayerNorm and Conditional Modulation}
Conditional modulation~\cite{huang2017arbitrary, perez2018film}, and in particular Adaptive LayerNorm~\cite{peebles2023scalable}, lets shared parameters behave differently under different conditions. It is now standard in diffusion transformers. Looped LLMs face an analogous need across iterations. LoopFormer~\cite{jeddi2026loopformer} conditions on the iteration index, but its condition vector is shared across all tokens in a pass. This is restrictive for language modeling where tokens at different positions warrant different updates~\cite{ainslie2023colt5, heakl2025drllm}. Our proposed IterAdaLN addresses this by conditioning on both the iteration and the per-token state (Section~\ref{sec:method-tadaln}).

%% file: sections/3-method.tex
\subsection{Overview}
\label{sec:method-overview}

\begin{figure}[t]
    \centering
    \includegraphics[width=0.95\columnwidth]{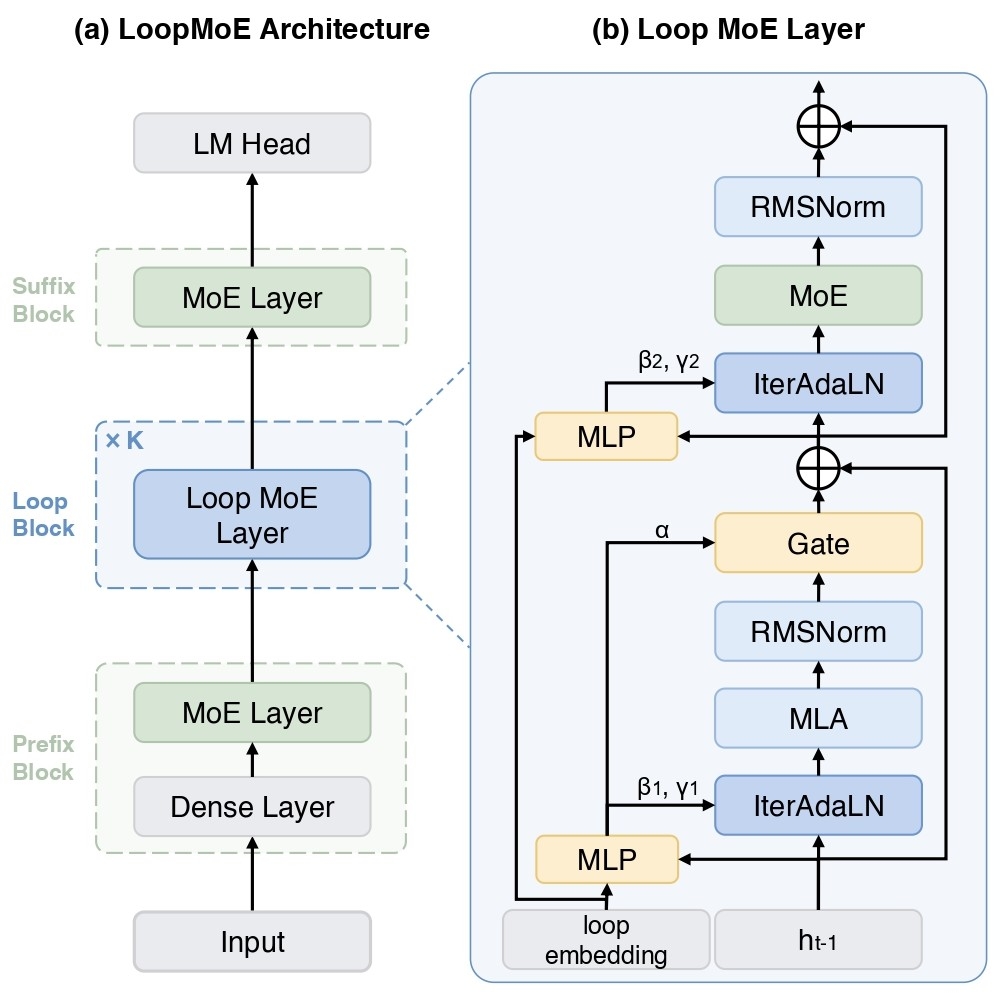}
    \caption{Architecture of \ourmodelnospace. \textbf{Left:} the sandwich-loop layout with prefix layers, a loop block of shared layers executed $K$ times, and suffix layers. \textbf{Right:} the per-iteration update inside the Loop MoE Layer, featuring IterAdaLN at both pre-normalization sites and an asymmetric residual gate $\alpha$ on the attention branch only.
    }
    \label{fig:model_arch}
\end{figure}

We propose \ourmodelnospace, a looped MoE language model. Following DeepSeek-V3~\cite{liu2024deepseek}, each layer pairs a Multi-head Latent Attention (MLA)~\cite{deepseekv2} sublayer with a fine-grained MoE sublayer, and the first post-embedding layer uses a dense FFN. Each sublayer adopts a sandwich normalization scheme, with pre-normalization before the sublayer and an additional RMSNorm on its output prior to the residual addition. On top of this backbone, we follow the sandwich-loop layout from prior weight-shared architectures~\cite{mcleish2025teaching,zeitoun2026hyperloop, geiping2026scaling}, organizing layers into non-shared prefix and suffix blocks around a shared loop body executed $K$ times. Inside the loop body, we introduce IterAdaLN (Section~\ref{sec:method-tadaln}) to supply per-iteration, per-token modulation, and a capacity-balancing strategy (Section~\ref{sec:method-budget}) to align total parameters, per-token FLOPs, and active sublayer ratios with a non-looped MoE reference of matched capacity (the Vanilla MoE). Figure~\ref{fig:model_arch} illustrates the overall layout and the per-iteration update inside one loop-block layer.

\subsection{Sandwich-Loop Backbone}
\label{sec:method-backbone}

\ourmodel organizes its layers into three regions, namely a non-shared prefix layers $L_p$, a loop block of $L_b$ shared layers executed $K$ times, and non-shared suffix layers $L_s$. This configuration yields an effective depth of $D = L_p + K \cdot L_b + L_s$ while using only $U = L_p + L_b + L_s$ unique layers. The non-shared boundary layers provide dedicated capacity for input-embedding adaptation and output projection, while grouping the shared layers into one multi-layer block enables implicit intra-block information exchange across iterations. In particular, a refined representation produced by a later layer in iteration $k$ becomes available to an earlier layer in iteration $k\!+\!1$, allowing the block as a whole to revisit and rewrite its intermediate states.


A central design question is how information flows across iterations. Prior sandwich-loop models~\cite{mcleish2025teaching,geiping2026scaling} re-inject the prefix output $h_{\text{pre}}$ additively at every step, requiring an ill-defined $h_0$ at the first iteration and allowing a constant $h_{\text{pre}}$ to dominate the residual stream, thereby suppressing iteration-to-iteration differentiation. We therefore remove the re-injection pathway so that the loop body takes $h_{\text{pre}}$ as its starting point and evolves through $K$ steps of pure recursion.
\begin{align}
\small
  h_0 &= h_{\text{pre}} \label{eq:loop-init} \\
  h_k &= f_{\mathrm{body}}(h_{k-1};\,\mathbf{c}_{k,t}), \quad k = 1, 2, \dots, K
  \label{eq:pure-recursion}
\end{align}
where $\mathbf{c}_{k,t}$ is the conditioning vector for token $t$ at step $k$ and is subsequently consumed directly by IterAdaLN (Section~\ref{sec:method-tadaln}). Each forward pass is therefore a strict recursive refinement, and IterAdaLN becomes the sole pathway for per-iteration differentiation.

\subsection{IterAdaLN}
\label{sec:method-tadaln}

Weight sharing constrains the model to reuse identical parameters across all steps, so the computation lacks iteration awareness without explicit conditioning. A natural remedy is to condition each iteration on a global iteration signal via Adaptive Layer Normalization (AdaLN)~\cite{peebles2023scalable}. This is sufficient for image generation, where the diffusion timestep serves as a genuinely global conditioning target. Applying the same recipe to an MoE language model backbone, however, introduces a spatial bottleneck. A purely iteration-conditioned AdaLN forces a uniform affine modulation across all sequence positions at each step $k$, whereas natural language sequences are intrinsically heterogeneous. Syntactic markers, function words, and dense semantic entities demand qualitatively different representational updates at the same iteration. With only a global signal available, all sequence-level heterogeneity must be absorbed by the weights of the shared body.

We therefore introduce \textbf{IterAdaLN}, which generates modulation parameters jointly from the iteration index and the current token state, unlocking per-iteration and per-token degrees of freedom directly within the conditioning pathway. For token position $t$ at loop iteration $k$, IterAdaLN forms branch-specific joint conditions $c^{\mathrm{attn}}_{k,t}$ and $c^{\mathrm{moe}}_{k,t}$ by fusing two streams of information. The token stream supplies local semantic context through a linear projection of the token's current state in the forward pass. For the attention branch, this state is the initial state $h_{k-1,t}$, while for the MoE branch, it is the intermediate state $m_{k,t}$ produced by the preceding attention update. The iteration stream follows the DiT-style time encoding~\cite{peebles2023scalable}, mapping the iteration index $k$ through a fixed sinusoidal positional encoding $\mathrm{PE}(k)$ and a small learnable MLP, $\mathbf{v}_k = \mathrm{MLP}_\mathrm{iter}(\mathrm{PE}(k))$. The two streams are combined by broadcast addition for each branch,

\begingroup
\small
\begin{align}
  c^{\mathrm{attn}}_{k,t} &= \mathrm{Linear}_{\mathrm{attn}}(h_{k-1,t})
              + \mathrm{MLP}_{\mathrm{iter}}\!\big(\mathrm{PE}(k)\big)
    \label{eq:c_kt_attn} \\
  c^{\mathrm{moe}}_{k,t} &= \mathrm{Linear}_{\mathrm{moe}}(m_{k,t})
              + \mathrm{MLP}_{\mathrm{iter}}\!\big(\mathrm{PE}(k)\big)
    \label{eq:c_kt_moe}
\end{align}
\endgroup
\noindent so that both conditions inherit sequence-length variation from the token term and iteration variation from the time term.

Given the corresponding condition $c$, IterAdaLN replaces standard RMSNorm~\cite{zhang2019root} at every pre-normalization site with

\begingroup
\small
\begin{equation}
  \mathrm{IterAdaLN}(h;\,c) =
     \frac{\big(1+\gamma(c)\big) \odot h}{\sqrt{\mathbb{E}[h^2]+\epsilon}}
    + \beta(c)
  \label{eq:iteradaln}
\end{equation}
\endgroup
\noindent Note that IterAdaLN employs an affine-free RMSNorm. We omit the default static learnable scale of RMSNorm, since IterAdaLN dynamically supplies a token- and iteration-conditional scale.

To provide fine-grained, independent modulation, the affine parameters are generated by two separate, zero-initialized MLPs,

\begingroup
\small
\begin{align}
  \big[\gamma^{\mathrm{attn}}_{k,t},\,\beta^{\mathrm{attn}}_{k,t},\,\alpha_{k,t}\big]
  &= \mathrm{MLP}^{\mathrm{attn}}_{\mathbf{0}}\!\big(c^{\mathrm{attn}}_{k,t}\big)
    \label{eq:mlp_attn} \\
  \big[\gamma^{\mathrm{moe}}_{k,t},\,\beta^{\mathrm{moe}}_{k,t}\big]
  &= \mathrm{MLP}^{\mathrm{moe}}_{\mathbf{0}}\!\big(c^{\mathrm{moe}}_{k,t}\big)
    \label{eq:mlp_moe}
\end{align}
\endgroup
\noindent where the subscript $\mathbf{0}$ denotes zero initialization of the output projection.

We apply the residual gate $\alpha_{k,t}$ only to the attention branch. A symmetric gate on the MoE branch would be absorbed into the routing weights and interfere with load-balancing calibration (see Appendix~\ref{app:alpha-asymmetric} for justification).



We accordingly express the per-token, per-iteration layer update as the sequential composition of an attention residual sub-block and an MoE residual sub-block. The intermediate state $m_{k,t}$ is first produced by the attention block,
\begin{equation}
\resizebox{0.95\linewidth}{!}{$
  m_{k,t} = h_{k-1,t} + \alpha_{k,t} \, \mathcal{A}\big(\mathrm{IterAdaLN}(h_{k-1,t}; \, \gamma^{\text{attn}}_{k,t}, \, \beta^{\text{attn}}_{k,t})\big)
$}
\label{eq:layer-update-attn}
\end{equation}
where $\mathcal{A}$ is the MLA attention sublayer. This updated state $m_{k,t}$ is then used both to compute the condition $c^{\mathrm{moe}}_{k,t}$ and as the input to the MoE block, yielding the final layer output,
\begin{equation}
\resizebox{0.95\linewidth}{!}{$
  h_{k,t} = m_{k,t} + \mathrm{MoE}\!\big(
         \mathrm{IterAdaLN}(m_{k,t};\, \gamma^{\mathrm{moe}}_{k,t},\, \beta^{\mathrm{moe}}_{k,t})
       \big)
$}
\label{eq:layer-update-attn}
\end{equation}
These update rules reflect the asymmetric placement of $\alpha_{k,t}$. Following AdaLN-Zero~\cite{peebles2023scalable}, the output projection of $\mathrm{MLP}_{\mathbf{0}}$ is zero-initialized, so that $\gamma_{k,t}=\beta_{k,t}=\alpha_{k,t}=0$ at initialization. The $K$-step loop therefore begins as a sequence of near-identity transformations, from which per-iteration and per-token differentiation gradually emerges during optimization. Appendix~\ref{app:loop-alpha-schedule} empirically verifies that the trained model exploits this per-iteration, per-token freedom, exhibiting a back-loaded, attention-concentrated update schedule.

\subsection{Capacity Balancing Strategy}
\label{sec:method-budget}

Looped MoEs exhibit a structural distortion in active capacity allocation. The dense MLA sublayer reuses the same parameters at every iteration, while a token may route to different experts across the $K$ iterations of a shared MoE layer, so that its unique active FFN parameters accumulate with $K$. The resulting ratio $\rho_{\text{loop}} = A_{\text{attn}}/A_{\text{ffn}}$ drops well below the operating point $\rho^\star$ of well-tuned non-looped MoEs, leaving attention under-parameterized relative to the FFN pool (Section~\ref{sec:ablation}). Let $N$ and $F$ denote a model's total parameter count and per-token FLOPs, with $N^\star, F^\star$ the corresponding baseline values. As noted in Section~\ref{sec:intro}, MoE sparsity decouples $N$ from $F$, so once $\rho$ is corrected we can independently restore $N$ and $F$ to $N^\star$ and $F^\star$. We achieve this alignment through a two-step procedure.

\paragraph{Measurement}
We first experimentally investigate the expected number of unique experts a token activates across the $K$ iterations of one shared layer under top-$k$ routing. The resulting $A_{\text{ffn}}$ scales sublinearly but non-trivially with $K$, while $A_{\text{attn}}$ remains constant (see Appendix~\ref{app:active-scaling}). We adopt a Vanilla MoE, a non-looped MoE, as our matched reference. We characterize it by its attention-to-FFN active ratio $\rho^\star = A^\star_{\text{attn}}/A^\star_{\text{ffn}}$, together with $N^\star$ and $F^\star$ introduced above, which jointly specify the target operating point.

\paragraph{Rebalancing and capacity restoration}
To elevate $\rho$ toward $\rho^\star$, we expand the MLA low-rank projections to increase $A_{\mathrm{attn}}$, and shrink each expert's hidden dimension to decrease $A_{\mathrm{ffn}}$, while keeping top-$k$ routing fixed. This preserves $F$ but reduces $N$. We then restore $N$ to $N^\star$ by scaling up the routed expert pool, exploiting the MoE decoupling of $N$ from $F$.

%% file: sections/4-experimental_setup.tex
\subsection{Data}
We pre-train all models on the publicly released OLMo-3 pre-training corpus Dolma3Mix~\citep{olmo2025olmo}, a large-scale mixture spanning web documents, code, academic text, and curated high-quality sources. To enable a strictly controlled comparison, all our 3B models consume an identical 200B-token subset of this corpus under the same document ordering and packing configuration, while the token budget for 9B models is 300B tokens.

\subsection{Training and Evaluation Setting}
All models are trained with the AdamW optimizer~\cite{loshchilov2017decoupled} under a cosine learning-rate schedule~\cite{loshchilov2017sgdr} with linear warmup. For MoE layers, we adopt top-$k$ routing with $k\ {=}\  6$ together with a shared expert strategy following~\cite{dai2024deepseekmoe}, and this routing configuration is identical between \ourmodel and Vanilla MoE. Our main 3B \ourmodel comprises $6$ unique layers, organized as $2$ prefix layers, a $2$-layer block that is recurrently applied for $4$ iterations with shared weights, and $2$ suffix layers, which yields an effective depth of $12$ layers. The 9B variants follow the same architectural recipe with proportionally scaled hidden dimensions and expert counts. Full hyperparameters for both scales are provided in Appendix~\ref{app:hyperparams}.

\begin{table*}[htbp]
\centering
\footnotesize
\setlength{\tabcolsep}{3.5pt}
\renewcommand{\arraystretch}{1.1}
\resizebox{\textwidth}{!}{%
\begin{tabular}{l c c cc ccc cc cc c}
\toprule
& \multicolumn{1}{c}{\textbf{Params}} & \textbf{Train} 
& \multicolumn{6}{c}{\textbf{Language \& Knowledge}} 
& \multicolumn{3}{c}{\textbf{Reasoning \& Math}} 
& \multirow{2}{*}{\textbf{Avg.}} \\
\cmidrule(lr){4-9} \cmidrule(lr){10-12}
\textbf{Model} & \textbf{(T/A)} & \textbf{Tokens} 
& MMLU & HSwag & PIQA & WGrd & TQA & RACE & BBH & GSM8K & MATH & \\
\midrule
\multicolumn{13}{l}{\textit{Dense Models}} \\
Pythia-1.4B$^{\dagger}$        & 1.4B / 1.4B & 200B   & 24.63 & 51.93 & 70.46 & 56.59 & 16.37 & 35.22& 23.71 & 1.90 & 1.80 & 31.30\\
OLMo2-1B$^{\dagger}$            & 1.0B / 1.0B & 200B   & 24.27 & 58.23 & 73.72 & 59.75 & 26.13 & 33.88& 9.45 & 2.43 & 0.42 & 32.12\\
Qwen-3-1.7B$^{\star}$           & 1.7B / 1.7B & 36T     & 62.46 & 67.09 & -- & 66.30 & -- & -- & 53.51 & 70.28 & 25.80 & -- \\
\midrule
\multicolumn{13}{l}{\textit{Dense Loop / Recursive Models}} \\
Hyperloop$^{\star}$       &   1B / 1B & 100B     & -- & 46.22 & 68.72 & 53.04 & -- & 33.59 & -- & -- & -- & -- \\
Griffin$^{\star}$             & 1B / 1B & 300B     & 29.5 & 67.2 & 77.4 & 65.2 & -- & -- &--& -- & -- & -- \\
Ouro-1.4B-R4$^{\star}$          & 1.4B / 1.4B & 7.7T   & 67.35 & 74.29 & -- & 72.30 & -- & -- & 71.02 & 78.92 & 82.40 & -- \\
\midrule
\multicolumn{13}{l}{\textit{MoE Models}} \\
DeepSeekMoE-2B$^{\star}$       & 2B / 0.3B & 100B   & -- & 54.8 & 72.3 & -- &16.6& -- & --& -- & -- & -- \\
OLMoE-1B-7B$^{\dagger}$        & 7B / 1B & 200B   & 24.90 & 68.14 & 76.88 & 60.62 & 39.28&35.31& 26.85& 2.82 & 1.66 & 37.38\\
PowerMoE-3B                & 3B / 0.8B  & 3T     & 43.24 & 73.07 & 79.00 & 64.80& 28.57&35.79& 33.07& 38.21 & 14.94 & 45.63\\
\midrule
\multicolumn{13}{l}{\textit{Ours}} \\
Vanilla MoE     & 3B / 0.8B    & 200B   & \textbf{27.86} & 58.68 & 74.12 & 49.92 & 36.19& 35.00& 26.12& 4.12 & 1.53 & 34.84\\
\textbf{\ourmodelnospace}         & 3B / 0.8B$^{\ddagger}$    & 200B   & 26.81 & \textbf{60.98} & \textbf{75.22} & \textbf{51.42} & \textbf{37.63} & \textbf{35.77} & \textbf{27.94} & \textbf{5.26} & \textbf{1.88} & \textbf{35.89} \\
\bottomrule
\end{tabular}}
\caption{
Main results across 9 downstream benchmarks. 
\textbf{(T/A)}: total / active parameters. 
$^{\dagger}$ evaluated at the 200B-token intermediate checkpoint for token-matched comparison.
$^{\star}$ numbers reported from the original paper.
$^{\ddagger}$ Active parameters are reported as 0.8B to match the FLOPs-per-token of the Vanilla MoE for fair comparison; the actual physical active parameter count of \ourmodel is 0.6B, since shared attention parameters across $K$ loop iterations contribute once to the physical count but $K$ times to per-token FLOPs.
HSwag stands for HellaSwag; WGrd stands for WinoGrande; TQA stands for TriviaQA.
Best results between Vanilla MoE and \ourmodel are in \textbf{bold}.
}
\label{tab:main_results}
\end{table*}

To assess downstream performance, we evaluate the models on a suite of $9$ standard benchmarks covering two broad dimensions. The first dimension targets knowledge and language understanding, including MMLU~\cite{hendryckstest2021}, HellaSwag~\cite{zellers2019hellaswag}, PIQA~\cite{bisk2020piqa}, WinoGrande~\cite{sakaguchi2021winogrande}, TriviaQA~\cite{joshi2017triviaqa}, and RACE~\cite{lai2017race}. The second dimension focuses on reasoning and mathematics, covering BBH~\cite{kazemi2025big}, GSM8K~\cite{cobbe2021training}, and MATH~\cite{hendrycksmath2021}. Additionally, we replace MMLU with the ARC datasets~\cite{allenai:arc} in the ablation for more precise analysis. All evaluations are conducted with lm-evaluation-harness~\cite{eval-harness} under the standard zero-shot or few-shot protocols prescribed for each benchmark. The evaluation settings are detailed in Appendix~\ref{app:lm_eval}. For external baselines whose full training exceeds 200B tokens, we report results at their 200B-token intermediate checkpoints when available, so that all numbers reflect a token-matched comparison.

%% file: sections/5-results.tex
\subsection{Main Results}
\label{sec:main_result}

Table~\ref{tab:main_results} compares \ourmodel against dense, dense-loop, and MoE baselines across nine benchmarks spanning knowledge, general language understanding, commonsense reasoning, multi-step reasoning, and mathematical problem solving. Vanilla MoE is the matched non-loop MoE described in Section~\ref{sec:method-budget}, sharing total parameters, per-token FLOPs, active sublayer ratio, and training tokens with \ourmodelnospace.

\ourmodel delivers consistent general improvements over the Vanilla MoE. It outperforms the baseline on 8 of the 9 benchmarks with an average gain of over 1 point, the only exception being a marginal regression on MMLU. These improvements span both evaluation dimensions, covering knowledge and language understanding as well as reasoning and mathematics. This shows that introducing the loop architecture into a modern MoE stack yields a broad net benefit under a strictly controlled compute budget. While the gains are broad, they are not uniform in magnitude. The largest improvements concentrate on reasoning and mathematics, where all datasets improve by a substantial margin relative to the average. We attribute this concentration to the additional iterative computation that the loop provides. Tasks that reward multi-step reasoning benefit most directly from repeated refinement of the hidden state. Section~\ref{sec:ablation} decomposes these contributions across the individual architectural components.

For broader context, Table~\ref{tab:main_results} also lists external systems trained on comparable token budgets (Pythia-1.4B~\cite{biderman2023pythia}, OLMo2-1B~\cite{olmo20242}, Hyperloop~\cite{zeitoun2026hyperloop}, DeepSeekMoE-2B~\cite{dai2024deepseekmoe}) and on substantially larger ones (Qwen-3-1.7B~\cite{yang2025qwen3}, Griffin~\cite{de2024griffin}, Ouro-1.4B-R4~\cite{zhu2025scaling}, PowerMoE-3B~\cite{shen2024power}). Against token-matched dense baselines, \ourmodel substantially outperforms them under identical training volume. On reasoning and mathematics, it further remains competitive with OLMoE-1B-7B~\cite{muennighoff2025olmoe}, despite using less than half the total parameters. We include the trillion-token systems only to mark the current performance frontier and do not treat them as head-to-head competitors, since their numbers conflate architecture with training scale and pipeline maturity. We regard large-scale training as complementary to the architectural axis isolated 
in this work.

\subsection{Scaling Behavior}
\label{sec:scaling}
\begin{table}[t]
\centering
\small
\begin{tabular}{lcc}
\toprule
\textbf{Model} & \textbf{Params (T/A)} & \textbf{Avg.} \\
\midrule
Vanilla MoE & 9B / 1.9B            & 41.22 \\
\textbf{\ourmodelnospace}     & 9B / 1.9B$^\ddagger$ & \textbf{44.03} \\
\bottomrule
\end{tabular}
\caption{Scaling results at 9B parameters under a token-matched training budget of 300B tokens. The average is computed over the same nine benchmarks as in Table~\ref{tab:main_results}; full per-benchmark results are provided in Appendix~\ref{app:scaling-full}. The actual physical active parameter count for $^\ddagger$ is 1.3B (see Appendix~\ref{app:hyperparams}).}
\label{tab:scaling}
\end{table}

\begin{table*}[ht!]
\centering
\small
\begin{tabular}{lccccccc}
\toprule
Architecture & ARC-E & ARC-C & HellaSwag & BBH & GSM8K & Avg.\\
\midrule
Vanilla MoE                  & 75.04 & \textbf{45.80} & 58.68 & 26.12 & 4.12 & 41.95\\
\midrule
Loop Base & 74.24 & 45.50 & 58.87 & 27.33 & 5.34 & 42.26\\
\quad + Balancing   & 75.04 & 44.86 & \textbf{61.32} & 27.58 & \textbf{5.49} & 42.86\\
\quad + iteration-only AdaLN   & 73.41 & 43.77 & 58.74 & 26.21 & 5.23 & 41.47\\
\quad + IterAdaLN   & 74.75 & 45.38 & 59.19 & 27.26 & 5.34 & 42.38\\
\ourmodelnospace\ (Ours)  & \textbf{75.21} & 45.72 & 60.98 & \textbf{27.94} & 5.26 & \textbf{43.02}\\
\bottomrule
\end{tabular}
\caption{Ablation study on the core components of the proposed architecture.}
\label{tab:ablation}
\end{table*}

A central question for looped architectures is whether their advantage persists as model capacity and training compute increase. To address this, we train 9B variants of both \ourmodel and the Vanilla MoE under the same training recipe and corpus, and report a strictly token-matched comparison at the training budget of 300B tokens. Table~\ref{tab:scaling} reports their performance across the same nine benchmarks used in Section~\ref{sec:main_result}. \ourmodel outperforms the matched Vanilla MoE by approximately 3 points on average, compared with a 1 point gap at the 3B scale. 

Two observations follow. First, the advantage of \ourmodel not only persists but becomes more pronounced in the larger-scale setting. Second, \ourmodel outperforms the Vanilla MoE on all nine benchmarks, with particularly large improvements on MMLU and GSM8K. The breadth and magnitude of these gains suggest that the iterative-depth advantage is not confined to the 3B regime and may strengthen as model capacity and training compute scale.

\subsection{Ablation Study}
\label{sec:ablation}

We dissect the contribution of each architectural component through a controlled ablation, presented in Table~\ref{tab:ablation}. We select five representative benchmarks that span the major capability dimensions considered in our evaluation: ARC-E and ARC-C (knowledge), HellaSwag (commonsense), BBH (multi-step reasoning) and GSM8K (math). We replace MMLU with ARC datasets in ablation beacause near-chance performance limit its diagnostic resolution for distinguishing architectural variants at this model and training scale. Starting from Loop Base, we independently evaluate capacity balancing, iteration-only AdaLN, IterAdaLN, and finally the combination of IterAdaLN with balancing to obtain the full LoopMoE model.

Moving from the Vanilla MoE to Loop Base introduces weight sharing across iterations. We observe clear gains on the reasoning and math benchmarks, together with mixed but mostly small changes on the knowledge and commonsense benchmarks. This pattern indicates that the loop architecture, even before any further conditioning or capacity balancing, already contributes most of the reasoning and mathematics improvement.

We next isolate the role of conditioning granularity. The iteration-only AdaLN variant follows the LoopFormer-style design, using a modulation signal shared across all tokens within each iteration, and underperforms Loop Base on all five benchmarks. IterAdaLN instead conditions jointly on the iteration index and per-token state, improving consistently over the iteration-only variant. These results show that iteration awareness alone is insufficient and that per-token conditioning is important for adaptive normalization in a looped MoE.

The final component, Balancing, restores the attention-to-FFN active-parameter ratio to that of a standard non-looped transformer, correcting the FFN-heavy skew induced by weight sharing (Section~\ref{sec:method}). Applied directly to Loop Base, Balancing substantially improves HellaSwag and yields additional gains on BBH, GSM8K, and ARC-E. When combined with IterAdaLN, it further improves HellaSwag, BBH, and both ARC benchmarks, accompanied by a small regression on GSM8K, consistent with prior observations that math relies more heavily on FFN computation~\cite{geva2021transformer, jin2025disentangling}. Overall, Balancing and IterAdaLN provide positive aggregate gains both independently and in combination, suggesting that they address complementary aspects of the looped architecture. The resulting LoopMoE model achieves the best overall performance among all ablated configurations.

%% file: sections/6-analysis.tex
\subsection{BBH Subtask Analysis}

\begin{table}[t]
\centering
\small
\setlength{\tabcolsep}{2.5pt}
\begin{tabular}{lcccc}
\toprule
\textbf{Task} & \textbf{Vanilla MoE} & \textbf{Loop Base} & \textbf{\ourmodelnospace}\\
\midrule
hyperbaton            & 31.25 & 50.83 & \textbf{51.25}\\
navigate              & 40.83 & 52.50 & \textbf{57.08}\\
salient trans. err. det. & \textbf{22.50} & 16.67 & 15.00\\
logical deduct. 5 obj. & \textbf{22.50} & 17.50 & 16.25 \\
\bottomrule
\end{tabular}
\caption{Representative BBH subtask scores for Vanilla MoE, Loop Base, and \ourmodelnospace. Here, ``salient trans. err. det.'' and ``logical deduct. 5 obj.'' stand for ``salient translation error detection'' and ``logical deduction five objects'', respectively. Full results across all 27 subtasks are provided in Appendix~\ref{appendix:bbh}.}
\label{tab:bbh_highlight}
\end{table}

The aggregate BBH gain of \ourmodel over the Vanilla MoE conceals a structured heterogeneity across subtasks. This offers finer-grained evidence on where iterative depth contributes most. Table~\ref{tab:bbh_highlight} reports the two subtasks with the largest gains and the two with the largest regressions.

The two largest improvements, hyperbaton ($+20.00$) and navigate ($+16.25$), both require answers to be constructed through a sequence of dependent reasoning steps in which each step builds on context accumulated from prior iterations. The Loop Base alone already accounts for the bulk of these gains, capturing nearly the entire improvement of hyperbaton and roughly three-quarters of the improvement of navigate. This pattern indicates that iterative depth is the dominant mechanism, while IterAdaLN and capacity balancing provide a smaller secondary contribution on tasks whose reasoning chains benefit from per-token modulation. The finding offers subtask-level corroboration of the main result that loop iterations supply the additional computation required for complex multi-step reasoning. 

The two largest regressions, salient translation error detection and five-object logical deduction, exhibit a monotonic decline that is already present in Loop Base. We attribute this to reduced per-step activation breadth, since matching the Vanilla MoE in compute forces LoopMoE to activate fewer physical parameters per iteration.

Across the full 27-subtask distribution, gains and regressions align consistently with this principle. \ourmodel improves on tasks where reasoning is compositional and benefits from iterative refinement and regresses on tasks whose correctness depends on broad, one-shot information access within each step.

\subsection{Iterative Routing Dynamics}
\label{sec:routing_analysis}

Beyond aggregate task performance, we next examine the iterative structure of a looped MoE block, which raises a natural question about expert allocation, namely, whether successive iterations activate similar or dissimilar expert subsets. Activating largely disjoint subsets would let each iteration perform a functionally distinct computation and effectively expand the active parameter budget, whereas converging to the same subset would form a stable expert coalition that applies a consistent transformation under repeated application. To probe which regime the loop operates in, we measure two pairwise similarities between iterations. Cosine similarity of router inputs tracks continuous representation drift, and Jaccard similarity of selected expert sets tracks discrete routing change.

\begin{figure}[t]
    \centering
    \includegraphics[width=0.48\columnwidth]{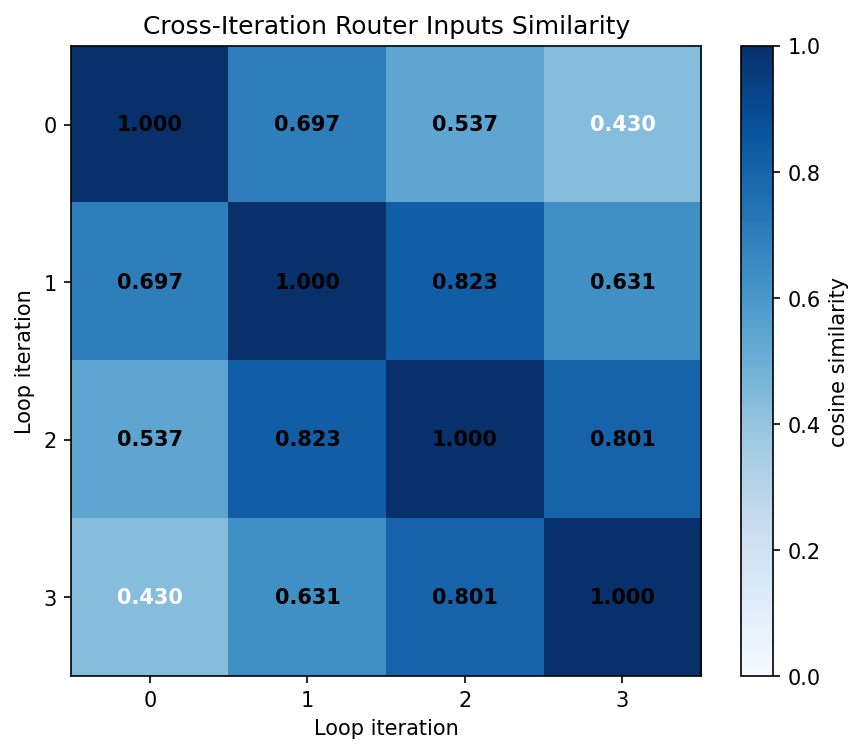}
    \includegraphics[width=0.48\columnwidth]{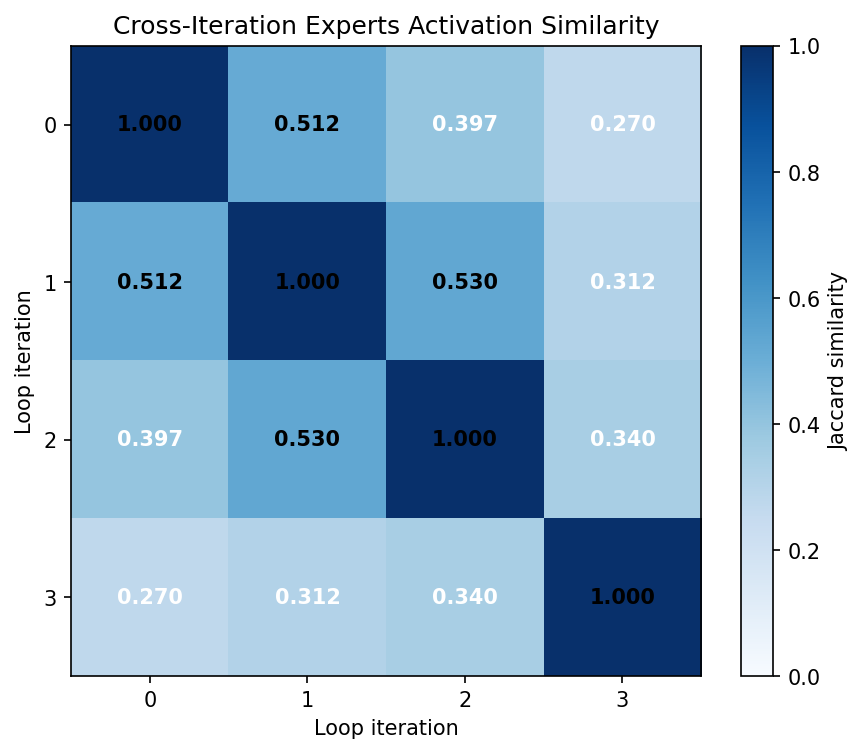}
    \caption{Cross-layer pairwise similarity between loop iterations on the BBH dataset. Router-input cosine (left) and expert-activation Jaccard (right). Per-layer matrices are reported in Appendix~\ref{app:per_layer_routing}.}
    \label{fig:loop_dynamics}
\end{figure}

The heatmaps in Figure~\ref{fig:loop_dynamics} reveal a three-phase structure that interpolates between the two extremes at different points in the loop. In the entry phase, iter~0 stands apart from all subsequent iterations and is especially distant in cosine similarity. This is the only iteration that reads directly from the non-loop prefix rather than from the previous block iteration, and it behaves as an adapter that projects the prefix distribution into the recurrent state. In the recurrent core, iters~1 and~2 form the tightest cluster and reach the highest pairwise Jaccard similarity, approaching the fixed-coalition regime. Once the prefix has been absorbed, the loop settles into a stationary internal recurrence carried by a consistent expert subset~\cite{blayney2026mechanistic}. In the exit phase, the terminal iteration exhibits a notable decoupling between the two metrics. Representations remain close to those of the previous iteration, yet the router selects a substantially different expert subset, which suggests that the final pass reformats the consolidated state for the downstream non-loop layers rather than continuing to refine it. A per-layer decomposition that shows how the two loop layers contribute asymmetrically to this structure is provided in Appendix~\ref{app:per_layer_routing}.

\subsection{Practical Efficiency}
Matching FLOPs controls the arithmetic cost of the compared models, but does not necessarily imply identical wall-clock efficiency. We therefore measure both training throughput and autoregressive decoding latency under identical hardware, parallelism, precision, and software configurations.

\paragraph{Training throughput.}
Table~\ref{tab:training_throughput} reports the achieved training throughput in TFLOP/s/GPU. Loop Base reaches 86.2 TFLOP/s/GPU, compared with 66.0 for Vanilla MoE, corresponding to a 31\% improvement. LoopMoE reaches 81.8 TFLOP/s/GPU, which is 24\% higher than the Vanilla MoE baseline, although it is 5\% lower than Loop Base due to
the additional components in LoopMoE. The looped variants benefit from fewer physical active parameters, which reduces communication latency and the time required for gradient updates.

\paragraph{Decoding latency.}
We further evaluate single-request decoding with batch size one at three context lengths. As shown in Table~\ref{tab:decoding_latency}, Loop Base is slightly faster than Vanilla MoE at all tested context lengths. Adding capacity balancing has negligible effect on latency, while adding IterAdaLN increases latency by approximately 9~ms/token relative to Loop Base. The looped architecture also leaves substantial room for further system-level optimization, which we leave for future work.

\begin{table}[t]
\centering
\small
\begin{tabular}{lc}
\toprule
Model & Throughput (TFLOP/s/GPU) \\
\midrule
Vanilla MoE             & 66.0 \\
Loop Base               & 86.2 \\
\textbf{LoopMoE (Ours)} & 81.8 \\
\bottomrule
\end{tabular}
\caption{
Achieved training throughput under identical hardware, parallelism, and training configurations.
}
\label{tab:training_throughput}
\end{table}

\begin{table}[t]
\centering
\small
\begin{tabular}{lccc}
\toprule
\multirow{2}{*}{Model}
& \multicolumn{3}{c}{Context Length} \\
\cmidrule(lr){2-4}
& 256 & 1024 & 2048 \\
\midrule
Vanilla MoE     & 66.495 & 65.759 & 64.919 \\
\midrule
Loop Base     & 64.321 & 64.173 & 64.012 \\
\quad + Balancing     & 63.809 & 64.026 & 64.223 \\
\quad + IterAdaLN     & 74.308 & 73.615 & 72.278 \\
\textbf{LoopMoE (Ours)}     & 72.461 & 72.292 & 71.042 \\
\bottomrule
\end{tabular}
\caption{
Autoregressive decoding latency (ms/token) under identical hardware and software configurations. 
}
\label{tab:decoding_latency}
\end{table}

%% file: sections/7-conclusion.tex

We introduced \ourmodelnospace, a looped MoE language model that couples sparse expert routing with iterative weight-shared computation. Two designs make this coupling viable. IterAdaLN breaks the symmetry imposed by weight sharing through a modulation signal jointly conditioned on the iteration index and the per-token hidden state. A capacity-balancing strategy recovers the attention-to-FFN active-parameter ratio of a well-tuned non-looped reference. Together, they enable the first rigorously matched evaluation of a looped architecture against a Vanilla MoE. Within this controlled setting, \ourmodel outperforms the Vanilla MoE on the vast majority of downstream tasks at the 3B scale while activating fewer physical parameters per token. At the 9B scale, \ourmodel outperforms its matched vanilla MoE on all benchmarks, with the average improvement increasing to nearly 3 points. These results provide evidence of positive scaling behavior and establish looped sparse computation as an effective architecture for improving overall model capability within a fixed capacity budget.

%% file: sections/limitations.tex
We acknowledge several limitations in our work. First, the dataset used for evaluation does not cover all domains, which may limit assessment of the model on specific or underrepresented fields. Second, our evaluation uses English as the main language, without consideration of multilingual scenarios. Third, while our results at 3B and 9B scales suggest that the architectural advantage is robust, confirming this trend at substantially larger scales remains an important direction for future work. All experiments are conducted within a limited training budget.

%% file: sections/appendix.tex
\subsection{Training Hyperparameters for \ourmodelnospace}
\label{app:hyperparams}
Table~\ref{tab:hyperparams} summarizes the architectural and training hyperparameters of LoopMoE and the Vanilla MoE. Values not explicitly differentiated are shared across the models to ensure a strictly controlled comparison. Due to the computational cost of pre-training 3B MoE models on 200B tokens and 9B MoE models on 300B tokens, all reported results correspond to a single training run, with the same random seed for all models.

\begin{table*}[h]
\centering
\small
\setlength{\tabcolsep}{3.5pt}
\renewcommand{\arraystretch}{1.05}
\begin{tabular}{lcccccc}
\toprule
& \multicolumn{3}{c}{\textbf{3B}} & \multicolumn{3}{c}{\textbf{9B}} \\
\cmidrule(lr){2-4} \cmidrule(lr){5-7}
\textbf{Hyperparameter} & \textbf{Vanilla MoE} & \textbf{Loop Base} & \textbf{LoopMoE} & \textbf{Vanilla MoE} & \textbf{Loop Base} & \textbf{LoopMoE} \\
\midrule
\multicolumn{5}{l}{\emph{Backbone}} \\
\midrule
Total parameters                 & 3B   & 3B   & 3B   & 9B   & 9B& 9B   \\
Active parameters                & 0.8B & 0.8B & 0.8B & 1.9B& 1.9B& 1.9B\\
Physical active parameters       & 0.8B & 0.6B & 0.6B & 1.9B& 1.3B& 1.3B\\
Effective depth                  & 12   & 12   & 12   & 18& 18& 18\\
Unique layers                    & 12   & 6    & 6    & 18& 9& 9\\
\quad Prefix layers $L_p$        & --   & 2    & 2    & --   & 3& 3\\
\quad Loop-block layers $L_b$    & --   & 2    & 2    & --    & 3& 3\\
\quad Suffix layers $L_s$        & --   & 2    & 2    & --   & 3& 3\\
Loop iterations $K$              & --   & 4    & 4    & --   & 4& 4\\
\midrule
\multicolumn{5}{l}{\emph{MLA}} \\
\midrule
Q LoRA Rank                      & 384  & 384  & 512  & 768& 768& 896\\
K/V LoRA Rank                    & 512  & 512  & 640  & 512& 512& 640\\
\midrule
\multicolumn{5}{l}{\emph{MoE-FFN}} \\
\midrule
routed experts                   & 64   & 128  & 136  & 64& 128& 136\\
shared experts                   & 2    & 2    & 2    & 2& 2& 2\\
Top-$k$                          & 6    & 6    & 6    & 6& 6& 6\\
FFN Hidden Size                  & 896  & 896  & 864  & 1280& 1280& 1216\\
\midrule
\multicolumn{5}{l}{\emph{Optimization}} \\
\midrule
Optimizer                        & \multicolumn{6}{c}{AdamW} \\
$(\beta_1, \beta_2)$             & \multicolumn{6}{c}{(0.9, 0.95)} \\
LR schedule                      & \multicolumn{6}{c}{cosine} \\
Warmup steps                     & \multicolumn{6}{c}{2000} \\
Peak learning rate               & 8.6e-4 & 8.6e-4& 8.6e-4 & 5.9e-4& 5.9e-4 & 5.9e-4 \\
Min.\ LR ratio                   & 7e-6 & 7e-6 & 7e-6 & 5e-6 & 5e-6& 5e-6\\
\bottomrule
\end{tabular}
\caption{Architectural and training hyperparameters of Vanilla MoE, Loop Base, and LoopMoE at 3B and 9B scales.}
\label{tab:hyperparams}
\end{table*}

\subsection{LM Evaluation Details}
We evaluate our models using the \texttt{lm-evaluation-harness} framework. To provide a clear overview of our evaluation protocol, the detailed settings for each downstream benchmark are summarized in Table \ref{tab:eval_details}.

\label{app:lm_eval}
\begin{table}[h]
\centering
\small
\begin{tabular}{lcc}
\toprule
\textbf{Benchmark} & \textbf{n-shot} & \textbf{Metric} \\
\midrule
MMLU       & 5-shot & Accuracy \\
HellaSwag  & 10-shot & Accuracy \\
PIQA       & 0-shot & Accuracy \\
WinoGrande & 0-shot & Accuracy \\
TriviaQA   & 5-shot & Exact Match \\
RACE       & 0-shot & Accuracy \\
BBH        & 3-shot & Exact Match \\
GSM8K      & 5-shot & Exact Match \\
MATH       & 4-shot & Exact Match \\
ARC-E       & 25-shot & Accuracy \\
ARC-C      & 25-shot & Accuracy \\
\bottomrule
\end{tabular}
\caption{Detailed evaluation settings across all downstream benchmarks.}
\label{tab:eval_details}
\end{table}

\subsection{Asymmetric Placement of the Residual Gate}
\label{app:alpha-asymmetric}
A critical design choice concerns where the residual gate $\alpha_{k,t}$ should act. While DiT applies a symmetric gate to both sublayer outputs, we apply $\alpha_{k,t}$ exclusively to the attention branch. This design follows directly from the structure of the MoE operator. For an IterAdaLN-modulated input $\tilde h_{k,t}$, the MoE output is a routing-weighted sum over selected experts, so that a branch-level residual gate is absorbed straight into those routing weights,

\begin{equation}
\small
  \alpha_{k,t}\!\cdot\!\mathrm{MoE}(\tilde h_{k,t})
  = \!\!\sum_{i\in\mathrm{Topk(r_{k,t})}}\!\!\big(\alpha_{k,t}\, w_{k,t,i}\big)\,
         E_i(\tilde h_{k,t}).
  \label{eq:gate-absorb}
\end{equation}
The router therefore already provides the same per-token reweighting that an MoE-branch gate would offer, so the gate adds no representational capacity. It also actively interferes with routing. The routing weights $w_{k,t,i}$ are calibrated by the load-balancing loss and the expert-bias update~\cite{liu2024deepseek} to remain in a well-behaved regime. Rescaling them by a data-dependent $\alpha_{k,t}$ perturbs that calibration along with the running statistics driving the expert-bias controller. The MLA attention sublayer, by contrast, has no analogous internal token-conditional reweighting mechanism, which makes a residual gate there strictly complementary.

\subsection{Active Parameter Scaling under Loop Iterations}
\label{app:active-scaling}

\begin{figure}[h]
\centering
\includegraphics[width=\linewidth]{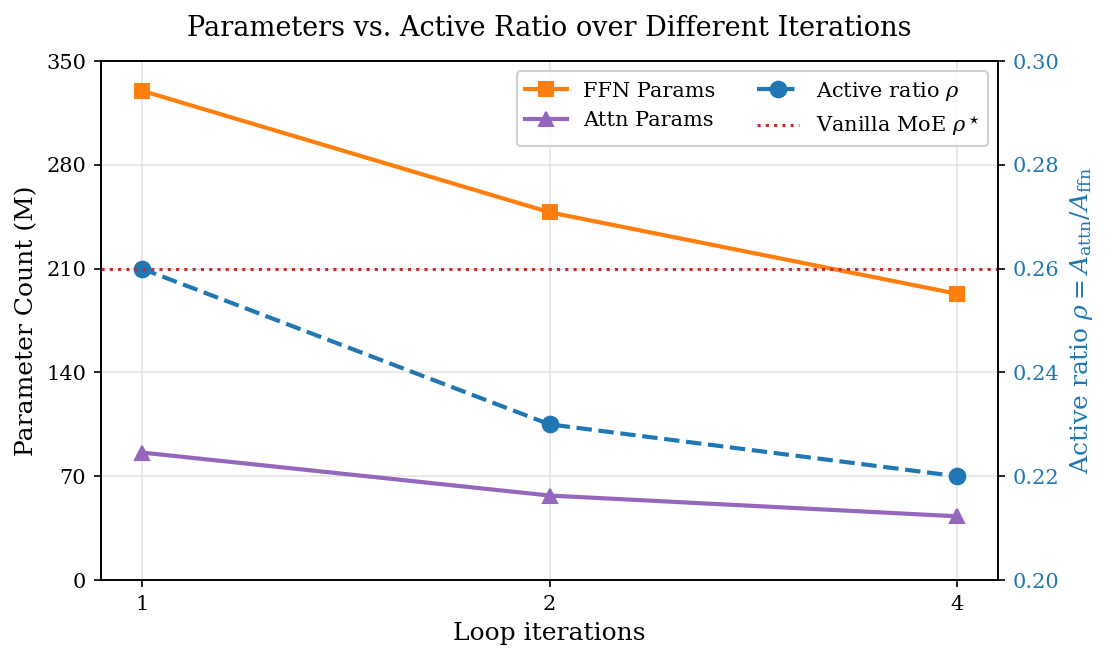}
\caption{Attention and FFN active parameters (left axis) and the active ratio $\rho = A_{\mathrm{attn}}/A_{\mathrm{ffn}}$ (right axis) as a function of loop iterations $K$. $A_{\mathrm{attn}}$ is constant in $K$ because the MLA sublayer is reused across iterations, while $A_{\mathrm{ffn}}$ grows sublinearly due to overlapping expert selections across iterations. The red dotted line marks the Vanilla MoE ratio $\rho^{\star}$.}
\label{fig:active-scaling}
\end{figure}

Figure~\ref{fig:active-scaling} empirically characterizes how the attention and FFN active parameters evolve as the number of loop iterations $K$ increases, together with the resulting attention-to-FFN active ratio $\rho = A_{\mathrm{attn}}/A_{\mathrm{ffn}}$. The dense MLA sublayer reuses the same parameters across all iterations, so $A_{\mathrm{attn}}$ remains constant in $K$. In contrast, a token may route to different experts at each iteration of the shared MoE layer, so its unique active FFN parameters $A_{\mathrm{ffn}}$ accumulate with $K$. The growth is, however, sublinear rather than $K\!\times\!$: as $K$ increases, the probability that a later iteration selects an already-activated expert grows, so the marginal contribution of each additional iteration to $A_{\mathrm{ffn}}$ shrinks. As a direct consequence, $\rho$ drops monotonically and falls well below the operating point $\rho^{\star}$ of well-tuned Vanilla MoE (red dotted line), motivating the capacity-balancing strategy described in Section~\ref{sec:method-budget}.

\subsection{Full 9B Scaling Results}
\label{app:scaling-full}

Table~\ref{tab:9b-full} reports the complete per-benchmark results for the 9B scaling comparison summarized in Section~\ref{sec:scaling}. Both LoopMoE and the Vanilla MoE are trained under identical recipes, corpus, and a strictly token-matched budget of 300B tokens, with all evaluation protocols identical to those in Section~\ref{sec:main_result}.

\begin{table*}[h]
\centering
\small
\setlength{\tabcolsep}{3.5pt}
\renewcommand{\arraystretch}{1.1}
\resizebox{\textwidth}{!}{%
\begin{tabular}{lcccccccccccc}
\toprule
\textbf{Model} & \textbf{Params (T/A)} & \textbf{MMLU} & \textbf{HSwag} & \textbf{PIQA} & \textbf{WGrd} & \textbf{TQA} & \textbf{RACE} & \textbf{BBH} & \textbf{GSM8K} & \textbf{MATH} & \textbf{Avg.} \\
\midrule
Vanilla MoE & 9B / 1.9B & 36.62 & 68.79 & 76.95 & 50.09 & 50.03 & 35.35 & 32.01 & 14.53 & 6.61 & 41.22 \\
Loop Base & 9B / 1.9B$^{\ddagger}$ & 45.28 & 69.77 & 76.92 & 50.32 & 50.81 & 35.64 & 32.17 & 17.07 & \textbf{8.32} & 42.92 \\
LoopMoE & 9B / 1.9B$^{\ddagger}$ & \textbf{45.33} & \textbf{70.24} & \textbf{77.51} & \textbf{51.91} & \textbf{52.90} & \textbf{37.40} & \textbf{32.89}& \textbf{20.94} & 7.12 & \textbf{44.03} \\
\bottomrule
\end{tabular}}
\caption{Full per-benchmark results at the 9B scale under a matched training budget of 300B tokens. The physical active parameter count of both $^{\ddagger}$-marked models is 1.3B, while the reported active parameter figure matches the per-token FLOPs of the Vanilla MoE for fair comparison. HSwag = HellaSwag, WGrd = WinoGrande, TQA = TriviaQA. Best results in each column are in \textbf{bold}.}
\label{tab:9b-full}
\end{table*}

The per-benchmark results confirm the pattern summarized in Section~\ref{sec:scaling}. \ourmodel outperforms the matched vanilla MoE on all nine benchmarks, with the largest absolute improvements appearing on MMLU and GSM8K. Its average advantage reaches 2.8 points, compared with 1.1 points at the 3B scale. Moreover, Loop Base already improves the average score by 1.7 points, showing that iterative computation provides substantial gains at 9B even before the complete \ourmodelnospace{} design is applied.

\subsection{BBH Detailed Subtask Results}
\label{appendix:bbh}
\begin{table*}[h]
\centering
\small
\setlength{\tabcolsep}{4pt}
\begin{tabular}{lccc}
\toprule
\textbf{Subtask} & \textbf{Vanilla MoE} & \textbf{Loop Base} & \textbf{\ourmodelnospace} \\
\midrule
boolean\_expressions                       & 49.58 & 52.08 & 54.58 \\
causal\_judgement                          & 47.16 & 48.86 & 50.57 \\
date\_understanding                        & 17.92 & 15.00 & 12.08 \\
disambiguation\_qa                         & 35.00 & 31.25 & 40.00 \\
dyck\_languages                            &  0.00 &  0.00 &  0.83 \\
formal\_fallacies                          & 41.25 & 53.33 & 48.33 \\
geometric\_shapes                          & 20.00 & 19.17 & 27.08 \\
hyperbaton                                 & 31.25 & 50.83 & 51.25 \\
logical\_deduction\_five\_objects          & 22.50 & 17.50 & 16.25 \\
logical\_deduction\_seven\_objects         & 14.17 & 13.33 & 13.33 \\
logical\_deduction\_three\_objects         & 34.17 & 26.25 & 32.50 \\
movie\_recommendation                      & 24.17 & 28.75 & 19.58 \\
multistep\_arithmetic\_two                 &  0.42 &  0.00 &  0.83 \\
navigate                                   & 40.83 & 52.50 & 57.08 \\
object\_counting                           & 17.08 & 28.33 & 22.92 \\
penguins\_in\_a\_table                     & 19.44 & 18.75 & 18.06 \\
reasoning\_about\_colored\_objects         & 12.08 & 15.83 &  9.58 \\
ruin\_names                                & 22.50 & 23.33 & 17.08 \\
salient\_translation\_error\_detection     & 22.50 & 16.67 & 15.00 \\
snarks                                     & 44.89 & 55.68 & 53.98 \\
sports\_understanding                      & 51.25 & 50.00 & 53.33 \\
temporal\_sequences                        & 25.00 & 26.67 & 25.42 \\
tracking\_shuffled\_objects\_five\_objects & 21.67 & 15.00 & 20.83 \\
tracking\_shuffled\_objects\_seven\_objects& 15.00 &  1.25 & 13.33 \\
tracking\_shuffled\_objects\_three\_objects& 32.50 & 32.92 & 33.75 \\
web\_of\_lies                              & 49.58 & 54.17 & 54.17 \\
word\_sorting                              &  1.25 &  0.42 &  1.67 \\
\midrule
\textbf{Average}                           & 26.12 & 27.33 & 27.94 \\
\bottomrule
\end{tabular}
\caption{Per-subtask scores on BBH for Vanilla MoE, Loop Base, and \ourmodelnospace.}
\label{tab:bbh-full}
\end{table*}

Table~\ref{tab:bbh-full} provides the complete per-task breakdown across all 27 subtasks within the BIG-bench Hard (BBH) suite. The disaggregated results provide a clearer view of the specific domains where the \ourmodel architecture excels.

\subsection{Per-layer Routing Dynamics in the Loop Body}
\label{app:per_layer_routing}
\begin{figure*}[t]
    \centering
    \includegraphics[width=0.98\columnwidth]{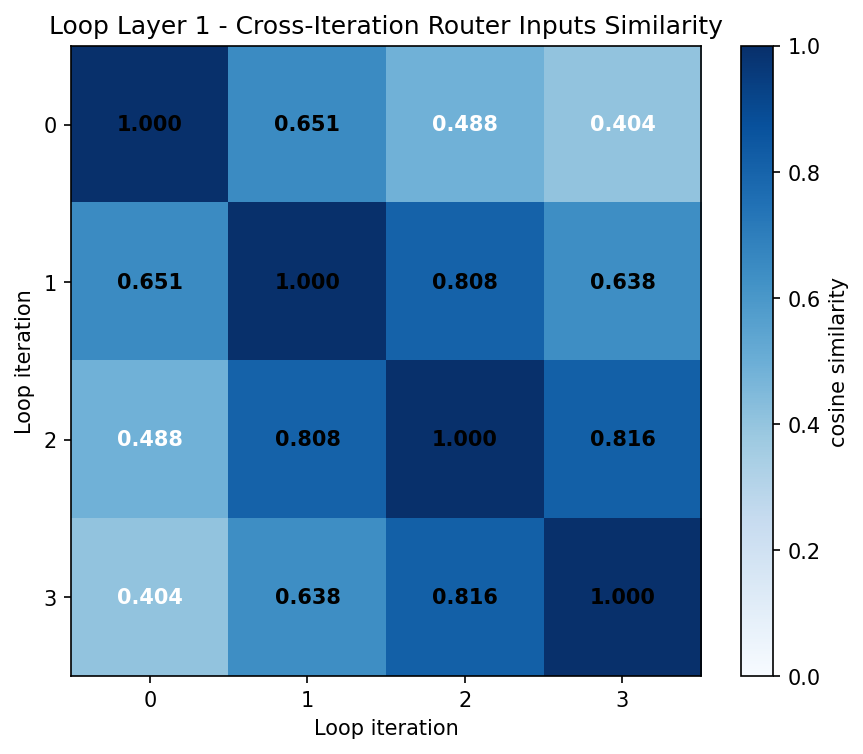}
    \includegraphics[width=0.98\columnwidth]{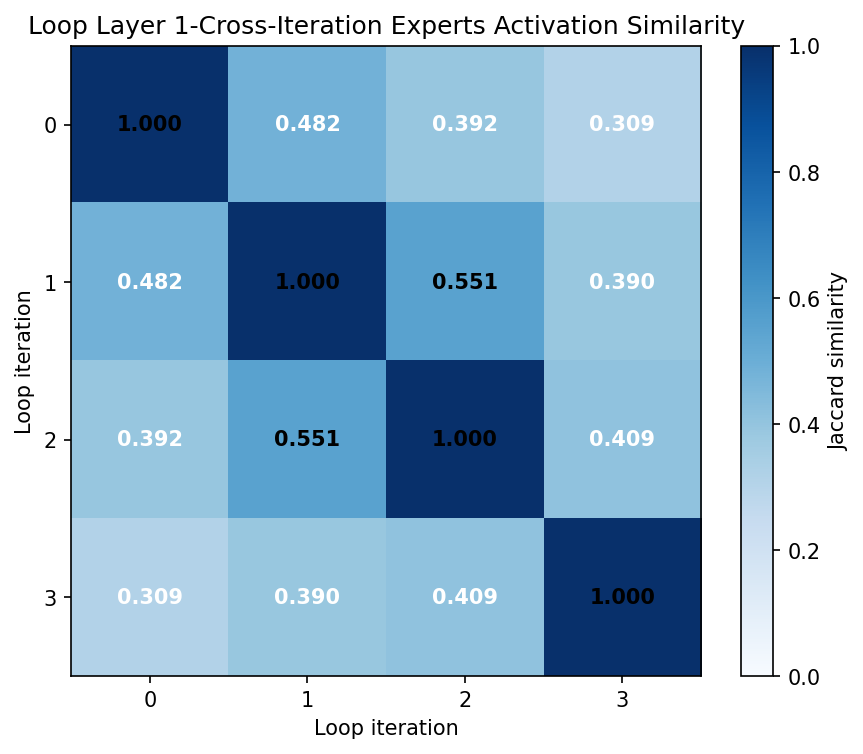}
    \includegraphics[width=0.98\columnwidth]{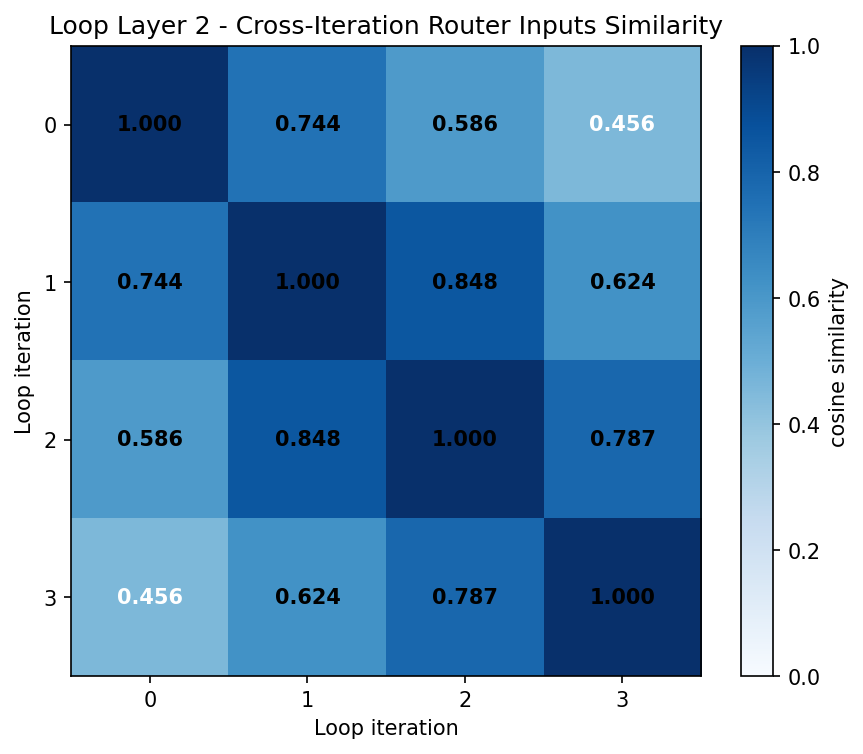}
    \includegraphics[width=0.98\columnwidth]{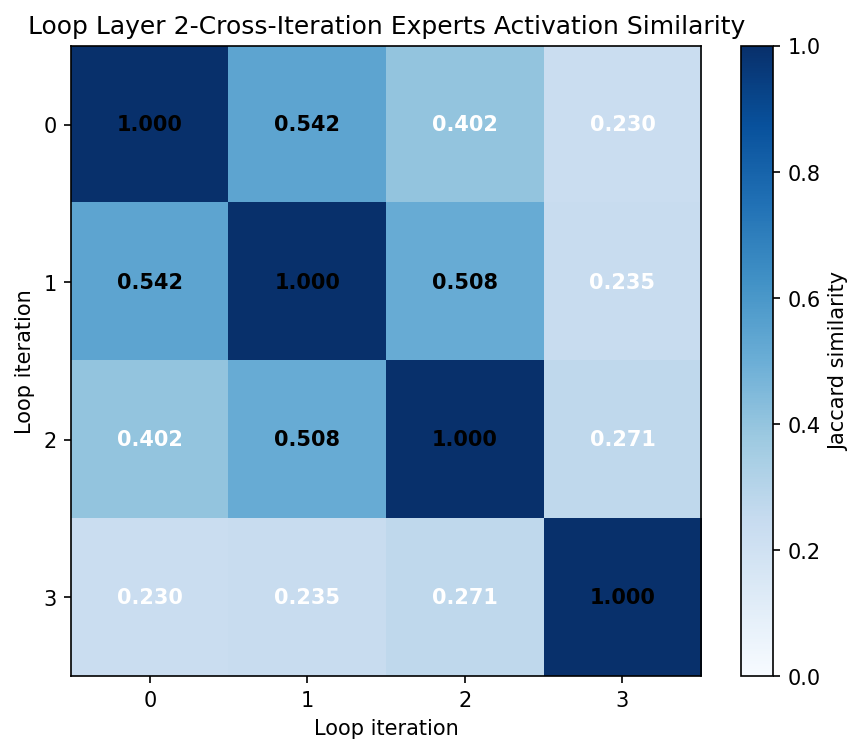}
    \caption{Per-layer cross-iteration routing dynamics within the shared block on the BBH dataset. The heatmaps display the router-input cosine similarity (left column) and expert-activation Jaccard similarity (right column) across the 4 loop iterations.}
    \label{fig:loop_dynamics_layers}
\end{figure*}
Figure~\ref{fig:loop_dynamics_layers} decomposes the three-phase routing structure into the two shared loop layers, revealing that they contribute asymmetrically.

\paragraph{Loop Layer~1} 
Loop Layer~1 shows monotonically increasing adjacent-pair cosine similarity across iterations, consistent with progressive fixed-point convergence of the recurrent block: once the prefix distribution has been absorbed at iter~0, successive applications drive the representation toward a stable attractor.

\paragraph{Loop Layer~2} 
Loop Layer~2 remains comparatively stable through the recurrent core but concentrates its functional shift at the read-out boundary, where its routing turnover (Jaccard drop at the terminal iteration) is substantially sharper than that of Loop Layer~1. This is consistent with Loop Layer~2 acting as the interface to the downstream non-loop suffix: rather than continuing to refine the recurrent state, it reformats it into a distribution suitable for the suffix layers.

\paragraph{Summary} 
Neither extreme intuition---fully disjoint per-iteration experts nor a single fixed coalition---holds in isolation. The loop adopts the fixed-coalition strategy within the recurrent core while employing distinct expert subsets at the two interface boundaries, with the entry layer (Loop Layer~1 at iter~0) absorbing the prefix and the exit layer (Loop Layer~2 at the terminal iteration) preparing the hand-off to the suffix. The entry and exit layers thus play complementary roles in mediating the transition between non-loop and recurrent computation.

\subsection{Emergent per-iteration allocation of residual updates}
\label{app:loop-alpha-schedule}

\begin{table*}[h]
\centering
\small
\setlength{\tabcolsep}{6pt}
\begin{tabular}{lcccccccc}
\toprule
Module & $i$ & $\overline{\gamma}$ & $\overline{|\gamma|}$ & $\overline{\beta}$ & $\overline{|\beta|}$ & $\overline{\alpha}$ & \textsc{$tok_{std}$} & \textsc{$tok_{range}$} \\
\midrule
\multirow{4}{*}{L1 attn} & 0 & $-0.159$ & $0.160$ & $-0.000$ & $0.001$ & $+0.707$ & $0.181$ & $0.805$ \\
                         & 1 & $+0.049$ & $0.171$ & $-0.000$ & $0.001$ & $+0.805$ & $0.227$ & $0.934$ \\
                         & 2 & $+0.237$ & $0.311$ & $-0.000$ & $0.001$ & $+1.044$ & $0.311$ & $1.273$ \\
                         & 3 & $+0.503$ & $0.545$ & $-0.000$ & $0.001$ & $+1.355$ & $0.486$ & $1.789$ \\
\midrule
\multirow{4}{*}{L1 ffn}  & 0 & $-0.339$ & $0.339$ & $-0.001$ & $0.001$ & --- & --- & --- \\
                         & 1 & $-0.276$ & $0.280$ & $-0.001$ & $0.001$ & --- & --- & --- \\
                         & 2 & $-0.196$ & $0.228$ & $-0.000$ & $0.001$ & --- & --- & --- \\
                         & 3 & $+0.008$ & $0.179$ & $+0.001$ & $0.002$ & --- & --- & --- \\
\midrule
\multirow{4}{*}{L2 attn} & 0 & $+0.034$ & $0.150$ & $+0.001$ & $0.001$ & $+0.601$ & $0.147$ & $0.761$ \\
                         & 1 & $+0.137$ & $0.227$ & $+0.002$ & $0.002$ & $+0.672$ & $0.225$ & $0.941$ \\
                         & 2 & $+0.253$ & $0.322$ & $+0.002$ & $0.002$ & $+0.839$ & $0.353$ & $1.373$ \\
                         & 3 & $+0.500$ & $0.554$ & $-0.001$ & $0.001$ & $+1.613$ & $0.922$ & $3.432$ \\
\midrule
\multirow{4}{*}{L2 ffn}  & 0 & $-0.252$ & $0.252$ & $+0.000$ & $0.000$ & --- & --- & --- \\
                         & 1 & $-0.222$ & $0.230$ & $+0.001$ & $0.001$ & --- & --- & --- \\
                         & 2 & $-0.168$ & $0.199$ & $+0.002$ & $0.002$ & --- & --- & --- \\
                         & 3 & $-0.075$ & $0.204$ & $+0.003$ & $0.003$ & --- & --- & --- \\
\bottomrule
\end{tabular}
\caption{Full per-iteration IterAdaLN modulation statistics for the two shared loop layers, averaged over the BBH dataset. $\overline{|\gamma|}$ and $\overline{|\beta|}$ denote the mean of absolute values of $\gamma$ and $\beta$, \texttt{$tok_{std}$} and \texttt{$tok_{rng}$} are the per-token standard deviation and range of $\alpha$.}
\label{tab:adaln-stats-full}
\end{table*}

\paragraph{Setup and observations.}
We instrument the IterAdaLN modulation outputs $(\gamma, \beta, \alpha)$ of the two shared loop layers (L1 and L2) at every iteration $i \in \{0,1,2,3\}$, separately for the attention branch and the FFN branch, averaged over the BBH dataset (full per-iteration statistics in Table~\ref{tab:adaln-stats-full}). Since the asymmetric residual gating design (Section~\ref{sec:method-tadaln}) fixes $\alpha\!=\!1$ on the FFN branch, $\alpha$ is not included in the ffn branch. Three findings emerge. First ~$\gamma$ increases monotonically across loop iterations in all four cells, but the endpoint regime differs by branch. Attention branches cross into amplification by the late loop, while FFN branches remain in mild contraction throughout (never crossing zero). Under the IterAdaLN input scaling, early iterations on every branch suppress the normalised input, while late iterations amplify it for attention but only relax the suppression for FFN. Second, attention $\alpha$ is strongly back-loaded. $\overline{\alpha}$ grows monotonically on both layers, with the final-iteration gate substantially exceeding the residual baseline of~1. The L2 endpoint of~1.613 means the last iteration's attention contribution is amplified by over 60\% relative to the residual baseline. Third, per-token dispersion of attention $\alpha$ grows monotonically with loop index. $tok_{std}(\alpha)$ increases roughly $3$--$6$ times, and the per-token range $tok_{range}(\alpha)$ on L2 attention grows from $0.761$ to $3.432$. This indicates that early iterations apply a nearly token-uniform update, and late iterations differentiate tokens strongly. The shift $\beta$ is negligible throughout ($|\overline{\beta}|<4\!\times\!10^{-3}$ across all cells).

\paragraph{Interpretation and implications}
We term this pattern a back-loaded update schedule with asymmetric branch roles. In the early loop iterations, both branches operate in a light-touch phase, characterised by input contraction, modest attention gating ($\alpha<1$), and near-token-uniform updates. The late iterations then diverge by branch. The attention branch enters a heavy-update phase, exhibiting input amplification ($\gamma>0$), strong gating, and strongly token-differentiated updates. The FFN branch, in contrast, implements persistent contraction, with $\gamma$ remaining $\leq 0$ at every iteration (with one marginal exception). Since FFN $\alpha\!\equiv\!1$ by design, the FFN contribution at every iteration is a consistently down-scaled update that merely becomes less down-scaled as the loop progresses. Effective late-loop computation is therefore concentrated specifically in the attention branch, both via its growing $\alpha$ gate and its growing per-token dispersion. This is consistent with emergent depth allocation under fixed-loop training. Weight sharing prevents the model from skipping iterations, but the model can learn to make early iterations approximate identity transformations and concentrate effective computation at the final iteration, choosing attention as the channel through which to do so. The monotonic growth of $tok_{std}(\alpha)$ and $tok_{range}(\alpha)$ reflects that per-token information about update magnitude is encoded precisely where differentiation is highest. Since $\gamma$ is conditioned on the evolving residual state $h$ rather than a directly zero-initialized scalar, the $\gamma$ sweep reflects learned behavior rather than initialization bias. This schedule does not yield a prefix-truncation inference speedup, since all four iterations are required to realise the late-loop attention amplification, and depth compressibility, to the extent it exists, is located at the front of the loop.

\subsection{Use of AI Assistants}
We used AI assistants for polishing the writing of this paper and for assisting with auxiliary coding tasks such as plotting and analysis scripts. All research ideas, experimental design, model implementation, and scientific claims are the authors' own, and all AI-generated content has been verified by the authors.

%% file: custom.bib
@article{liu2024deepseek,
  title={Deepseek-v3 technical report},
  author={Liu, Aixin and Feng, Bei and Xue, Bing and Wang, Bingxuan and Wu, Bochao and Lu, Chengda and Zhao, Chenggang and Deng, Chengqi and Zhang, Chenyu and Ruan, Chong and others},
  journal={arXiv preprint arXiv:2412.19437},
  year={2024}
}

@article{yang2025qwen3,
  title={Qwen3 technical report},
  author={Yang, An and Li, Anfeng and Yang, Baosong and Zhang, Beichen and Hui, Binyuan and Zheng, Bo and Yu, Bowen and Gao, Chang and Huang, Chengen and Lv, Chenxu and others},
  journal={arXiv preprint arXiv:2505.09388},
  year={2025}
}

@article{zeng2025glm,
  title={Glm-4.5: Agentic, reasoning, and coding (arc) foundation models},
  author={Zeng, Aohan and Lv, Xin and Zheng, Qinkai and Hou, Zhenyu and Chen, Bin and Xie, Chengxing and Wang, Cunxiang and Yin, Da and Zeng, Hao and Zhang, Jiajie and others},
  journal={arXiv preprint arXiv:2508.06471},
  year={2025}
}

@article{dehghani2018universal,
  title={Universal transformers},
  author={Dehghani, Mostafa and Gouws, Stephan and Vinyals, Oriol and Uszkoreit, Jakob and Kaiser, {\L}ukasz},
  journal={arXiv preprint arXiv:1807.03819},
  year={2018}
}

@article{lan2019albert,
  title={Albert: A lite bert for self-supervised learning of language representations},
  author={Lan, Zhenzhong and Chen, Mingda and Goodman, Sebastian and Gimpel, Kevin and Sharma, Piyush and Soricut, Radu},
  journal={arXiv preprint arXiv:1909.11942},
  year={2019}
}

@article{bae2026mixture,
  title={Mixture-of-recursions: Learning dynamic recursive depths for adaptive token-level computation},
  author={Bae, Sangmin and Kim, Yujin and Bayat, Reza and Kim, Sungnyun and Ha, Jiyoun and Schuster, Tal and Fisch, Adam and Harutyunyan, Hrayr and Ji, Ziwei and Courville, Aaron and others},
  journal={Advances in Neural Information Processing Systems},
  volume={38},
  pages={96572--96617},
  year={2026}
}

@inproceedings{bae2025relaxed,
  title={Relaxed recursive transformers: Effective parameter sharing with layer-wise lora},
  author={Bae, Sangmin and Fisch, Adam and Harutyunyan, Hrayr and Ji, Ziwei and Kim, Seungyeon and Schuster, Tal},
  booktitle={International Conference on Learning Representations},
  volume={2025},
  pages={34282--34327},
  year={2025}
}

@inproceedings{mohtashami2025cotformer,
  title={CoTFormer: A chain of thought driven architecture with budget-adaptive computation cost at inference},
  author={Mohtashami, Amirkeivan and Pagliardini, Matteo and Jaggi, Martin},
  booktitle={International Conference on Learning Representations},
  volume={2025},
  pages={11503--11520},
  year={2025}
}

@article{zeitoun2026hyperloop,
  title={Hyperloop transformers},
  author={Zeitoun, Abbas and Torroba-Hennigen, Lucas and Kim, Yoon},
  journal={arXiv preprint arXiv:2604.21254},
  year={2026}
}

@inproceedings{jeddi2026loopformer,
  title     = {LoopFormer: Elastic-Depth Looped Transformers for Latent Reasoning via Shortcut Modulation},
  author    = {Jeddi, Ahmadreza and Ciccone, Marco and Taati, Babak},
  booktitle = {International Conference on Learning Representations (ICLR)},
  year      = {2026},
  url       = {https://openreview.net/forum?id=RzYXb5YWBs}
}

@article{zhu2025scaling,
  title={Scaling latent reasoning via looped language models},
  author={Zhu, Rui-Jie and Wang, Zixuan and Hua, Kai and Zhang, Tianyu and Li, Ziniu and Que, Haoran and Wei, Boyi and Wen, Zixin and Yin, Fan and Xing, He and others},
  journal={arXiv preprint arXiv:2510.25741},
  year={2025}
}

@article{cai2025survey,
  title={A survey on mixture of experts in large language models},
  author={Cai, Weilin and Jiang, Juyong and Wang, Fan and Tang, Jing and Kim, Sunghun and Huang, Jiayi},
  journal={IEEE Transactions on Knowledge and Data Engineering},
  year={2025},
  publisher={IEEE}
}

@article{zhang2019root,
  title={Root mean square layer normalization},
  author={Zhang, Biao and Sennrich, Rico},
  journal={Advances in neural information processing systems},
  volume={32},
  year={2019}
}

@inproceedings{peebles2023scalable,
  title={Scalable Diffusion Models with Transformers},
  author={Peebles, William and Xie, Saining},
  booktitle={2023 IEEE/CVF International Conference on Computer Vision (ICCV)},
  pages={4172--4182},
  year={2023},
  organization={IEEE}
}

@article{lepikhin2020gshard,
  title={Gshard: Scaling giant models with conditional computation and automatic sharding},
  author={Lepikhin, Dmitry and Lee, HyoukJoong and Xu, Yuanzhong and Chen, Dehao and Firat, Orhan and Huang, Yanping and Krikun, Maxim and Shazeer, Noam and Chen, Zhifeng},
  journal={arXiv preprint arXiv:2006.16668},
  year={2020}
}

@article{fedus2022switch,
  title={Switch transformers: Scaling to trillion parameter models with simple and efficient sparsity},
  author={Fedus, William and Zoph, Barret and Shazeer, Noam},
  journal={Journal of Machine Learning Research},
  volume={23},
  number={120},
  pages={1--39},
  year={2022}
}

@article{hutchins2022block,
  title={Block-recurrent transformers},
  author={Hutchins, DeLesley and Schlag, Imanol and Wu, Yuhuai and Dyer, Ethan and Neyshabur, Behnam},
  journal={Advances in neural information processing systems},
  volume={35},
  pages={33248--33261},
  year={2022}
}

@article{geiping2026scaling,
  title={Scaling up test-time compute with latent reasoning: A recurrent depth approach},
  author={Geiping, Jonas and McLeish, Sean and Jain, Neel and Kirchenbauer, John and Singh, Siddharth and Bartoldson, Brian and Kailkhura, Bhavya and Bhatele, Abhinav and Goldstein, Tom},
  journal={Advances in Neural Information Processing Systems},
  volume={38},
  pages={41340--41391},
  year={2026}
}

@article{mcleish2025teaching,
    title={Teaching Pretrained Language Models to Think Deeper with Retrofitted Recurrence}, 
    author={Sean McLeish and Ang Li and John Kirchenbauer and Dayal Singh Kalra and Brian R. Bartoldson and Bhavya Kailkhura and Avi Schwarzschild and Jonas Geiping and Tom Goldstein and Micah Goldblum},
    journal={arXiv preprint arXiv:2511.07384},
    year={2025}
}

@inproceedings{perez2018film,
  title={Film: Visual reasoning with a general conditioning layer},
  author={Perez, Ethan and Strub, Florian and De Vries, Harm and Dumoulin, Vincent and Courville, Aaron},
  booktitle={Proceedings of the AAAI conference on artificial intelligence},
  volume={32},
  number={1},
  year={2018}
}

@inproceedings{huang2017arbitrary,
  title={Arbitrary style transfer in real-time with adaptive instance normalization},
  author={Huang, Xun and Belongie, Serge},
  booktitle={Proceedings of the IEEE international conference on computer vision},
  pages={1501--1510},
  year={2017}
}

@article{olmo2025olmo,
  title={Olmo 3},
  author={Olmo, Team and Ettinger, Allyson and Bertsch, Amanda and Kuehl, Bailey and Graham, David and Heineman, David and Groeneveld, Dirk and Brahman, Faeze and Timbers, Finbarr and Ivison, Hamish and others},
  journal={arXiv preprint arXiv:2512.13961},
  year={2025}
}

@article{loshchilov2017decoupled,
  title={Decoupled weight decay regularization},
  author={Loshchilov, Ilya and Hutter, Frank},
  journal={arXiv preprint arXiv:1711.05101},
  year={2017}
}

@inproceedings{loshchilov2017sgdr,
  title={SGDR: Stochastic Gradient Descent with Warm Restarts},
  author={Loshchilov, Ilya and Hutter, Frank},
  booktitle={International Conference on Learning Representations},
  year={2017}
}

@article{hendryckstest2021,
      title={Measuring Massive Multitask Language Understanding},
      author={Dan Hendrycks and Collin Burns and Steven Basart and Andy Zou and Mantas Mazeika and Dawn Song and Jacob Steinhardt},
      journal={Proceedings of the International Conference on Learning Representations (ICLR)},
      year={2021}
    }

@inproceedings{zellers2019hellaswag,
  title={Hellaswag: Can a machine really finish your sentence?},
  author={Zellers, Rowan and Holtzman, Ari and Bisk, Yonatan and Farhadi, Ali and Choi, Yejin},
  booktitle={Proceedings of the 57th annual meeting of the association for computational linguistics},
  pages={4791--4800},
  year={2019}
}

@inproceedings{bisk2020piqa,
  title={Piqa: Reasoning about physical commonsense in natural language},
  author={Bisk, Yonatan and Zellers, Rowan and Gao, Jianfeng and Choi, Yejin and others},
  booktitle={Proceedings of the AAAI conference on artificial intelligence},
  volume={34},
  number={05},
  pages={7432--7439},
  year={2020}
}

@article{sakaguchi2021winogrande,
  title={Winogrande: An adversarial winograd schema challenge at scale},
  author={Sakaguchi, Keisuke and Bras, Ronan Le and Bhagavatula, Chandra and Choi, Yejin},
  journal={Communications of the ACM},
  volume={64},
  number={9},
  pages={99--106},
  year={2021},
  publisher={ACM New York, NY, USA}
}

@inproceedings{joshi2017triviaqa,
  title={Triviaqa: A large scale distantly supervised challenge dataset for reading comprehension},
  author={Joshi, Mandar and Choi, Eunsol and Weld, Daniel S and Zettlemoyer, Luke},
  booktitle={Proceedings of the 55th Annual Meeting of the Association for Computational Linguistics (Volume 1: Long Papers)},
  pages={1601--1611},
  year={2017}
}

@inproceedings{lai2017race,
  title={Race: Large-scale reading comprehension dataset from examinations},
  author={Lai, Guokun and Xie, Qizhe and Liu, Hanxiao and Yang, Yiming and Hovy, Eduard},
  booktitle={Proceedings of the 2017 conference on empirical methods in natural language processing},
  pages={785--794},
  year={2017}
}

@inproceedings{kazemi2025big,
  title={Big-bench extra hard},
  author={Kazemi, Mehran and Fatemi, Bahare and Bansal, Hritik and Palowitch, John and Anastasiou, Chrysovalantis and Mehta, Sanket Vaibhav and Jain, Lalit K and Aglietti, Virginia and Jindal, Disha and Chen, Yuanzhu Peter and others},
  booktitle={Proceedings of the 63rd Annual Meeting of the Association for Computational Linguistics (Volume 1: Long Papers)},
  pages={26473--26501},
  year={2025}
}

@article{cobbe2021training,
  title={Training verifiers to solve math word problems},
  author={Cobbe, Karl and Kosaraju, Vineet and Bavarian, Mohammad and Chen, Mark and Jun, Heewoo and Kaiser, Lukasz and Plappert, Matthias and Tworek, Jerry and Hilton, Jacob and Nakano, Reiichiro and others},
  journal={arXiv preprint arXiv:2110.14168},
  year={2021}
}

@article{hendrycksmath2021,
  title={Measuring Mathematical Problem Solving With the MATH Dataset},
  author={Dan Hendrycks and Collin Burns and Saurav Kadavath and Akul Arora and Steven Basart and Eric Tang and Dawn Song and Jacob Steinhardt},
  journal={NeurIPS},
  year={2021}
}

@misc{eval-harness,
  author       = {Gao, Leo and Tow, Jonathan and Abbasi, Baber and Biderman, Stella and Black, Sid and DiPofi, Anthony and Foster, Charles and Golding, Laurence and Hsu, Jeffrey and Le Noac'h, Alain and Li, Haonan and McDonell, Kyle and Muennighoff, Niklas and Ociepa, Chris and Phang, Jason and Reynolds, Laria and Schoelkopf, Hailey and Skowron, Aviya and Sutawika, Lintang and Tang, Eric and Thite, Anish and Wang, Ben and Wang, Kevin and Zou, Andy},
  title        = {The Language Model Evaluation Harness},
  month        = 07,
  year         = 2024,
  publisher    = {Zenodo},
  version      = {v0.4.3},
  doi          = {10.5281/zenodo.12608602},
  url          = {https://zenodo.org/records/12608602}
}

@inproceedings{dai2024deepseekmoe,
  title={Deepseekmoe: Towards ultimate expert specialization in mixture-of-experts language models},
  author={Dai, Damai and Deng, Chengqi and Zhao, Chenggang and Xu, RX and Gao, Huazuo and Chen, Deli and Li, Jiashi and Zeng, Wangding and Yu, Xingkai and Wu, Yu and others},
  booktitle={Proceedings of the 62nd Annual Meeting of the Association for Computational Linguistics (Volume 1: Long Papers)},
  pages={1280--1297},
  year={2024}
}

@inproceedings{biderman2023pythia,
  title={Pythia: A suite for analyzing large language models across training and scaling},
  author={Biderman, Stella and Schoelkopf, Hailey and Anthony, Quentin Gregory and Bradley, Herbie and O’Brien, Kyle and Hallahan, Eric and Khan, Mohammad Aflah and Purohit, Shivanshu and Prashanth, USVSN Sai and Raff, Edward and others},
  booktitle={International Conference on Machine Learning},
  pages={2397--2430},
  year={2023},
  organization={PMLR}
}

@article{olmo20242,
  title={2 OLMo 2 Furious},
  author={OLMo, Team and Walsh, Pete and Soldaini, Luca and Groeneveld, Dirk and Lo, Kyle and Arora, Shane and Bhagia, Akshita and Gu, Yuling and Huang, Shengyi and Jordan, Matt and others},
  journal={arXiv preprint arXiv:2501.00656},
  year={2024}
}

@inproceedings{muennighoff2025olmoe,
  title={Olmoe: Open mixture-of-experts language models},
  author={Muennighoff, Niklas and Soldaini, Luca and Groeneveld, Dirk and Lo, Kyle and Morrison, Jacob and Min, Sewon and Shi, Weijia and Walsh, Pete and Tafjord, Oyvind and Lambert, Nathan and others},
  booktitle={International Conference on Learning Representations},
  volume={2025},
  pages={62061--62121},
  year={2025}
}

@article{de2024griffin,
  title={Griffin: Mixing gated linear recurrences with local attention for efficient language models},
  author={De, Soham and Smith, Samuel L and Fernando, Anushan and Botev, Aleksandar and Cristian-Muraru, George and Gu, Albert and Haroun, Ruba and Berrada, Leonard and Chen, Yutian and Srinivasan, Srivatsan and others},
  journal={arXiv preprint arXiv:2402.19427},
  year={2024}
}

@article{shen2024power,
  title={Power scheduler: A batch size and token number agnostic learning rate scheduler},
  author={Shen, Yikang and Stallone, Matthew and Mishra, Mayank and Zhang, Gaoyuan and Tan, Shawn and Prasad, Aditya and Soria, Adriana Meza and Cox, David D and Panda, Rameswar},
  journal={arXiv preprint arXiv:2408.13359},
  year={2024}
}

@inproceedings{geva2021transformer,
  title={Transformer feed-forward layers are key-value memories},
  author={Geva, Mor and Schuster, Roei and Berant, Jonathan and Levy, Omer},
  booktitle={Proceedings of the 2021 Conference on Empirical Methods in Natural Language Processing},
  pages={5484--5495},
  year={2021}
}

@inproceedings{jin2025disentangling,
  title={Disentangling memory and reasoning ability in large language models},
  author={Jin, Mingyu and Luo, Weidi and Cheng, Sitao and Wang, Xinyi and Hua, Wenyue and Tang, Ruixiang and Wang, William Yang and Zhang, Yongfeng},
  booktitle={Proceedings of the 63rd Annual Meeting of the Association for Computational Linguistics (Volume 1: Long Papers)},
  pages={1681--1701},
  year={2025}
}

@article{blayney2026mechanistic,
  title={A mechanistic analysis of looped reasoning language models},
  author={Blayney, Hugh and Arroyo, {\'A}lvaro and Obando-Ceron, Johan and Castro, Pablo Samuel and Courville, Aaron and Bronstein, Michael M and Dong, Xiaowen},
  journal={arXiv preprint arXiv:2604.11791},
  year={2026}
}

@misc{deepseekv2,
      title={DeepSeek-V2: A Strong, Economical, and Efficient Mixture-of-Experts Language Model}, 
      author={DeepSeek-AI},
      year={2024},
      eprint={2405.04434},
      archivePrefix={arXiv},
      primaryClass={cs.CL}
}

@article{hao2024training,
  title={Training large language models to reason in a continuous latent space},
  author={Hao, Shibo and Sukhbaatar, Sainbayar and Su, DiJia and Li, Xian and Hu, Zhiting and Weston, Jason and Tian, Yuandong},
  journal={arXiv preprint arXiv:2412.06769},
  year={2024}
}

@inproceedings{saunshi2025reasoning,
  title={Reasoning with latent thoughts: On the power of looped transformers},
  author={Saunshi, Nikunj and Dikkala, Nishanth and Li, Zhiyuan and Kumar, Sanjiv and J Reddi, Sashank},
  booktitle={International Conference on Learning Representations},
  volume={2025},
  pages={14855--14881},
  year={2025}
}

@inproceedings{ainslie2023colt5,
  title={Colt5: Faster long-range transformers with conditional computation},
  author={Ainslie, Joshua and Lei, Tao and de Jong, Michiel and Onta{\~n}{\'o}n, Santiago and Brahma, Siddhartha and Zemlyanskiy, Yury and Uthus, David C and Guo, Mandy and Lee-Thorp, James and Tay, Yi and others},
  booktitle={Proceedings of the 2023 conference on empirical methods in natural language processing},
  pages={5085--5100},
  year={2023}
}

@article{heakl2025drllm,
  title={Dr.LLM: Dynamic Layer Routing in LLMs},
  author={Ahmed Heakl and Martin Gubri and Salman Khan and Sangdoo Yun and Seong Joon Oh},
  journal={arXiv preprint arXiv:2510.12773},
  year={2025}
}

@article{jiang2024mixtral,
  title={Mixtral of experts},
  author={Jiang, Albert Q and Sablayrolles, Alexandre and Roux, Antoine and Mensch, Arthur and Savary, Blanche and Bamford, Chris and Chaplot, Devendra Singh and Casas, Diego de las and Hanna, Emma Bou and Bressand, Florian and others},
  journal={arXiv preprint arXiv:2401.04088},
  year={2024}
}

@article{xue2024openmoe,
  title={OpenMoE: An Early Effort on Open Mixture-of-Experts Language Models},
  author={Xue, Fuzhao and Zheng, Zian and Fu, Yao and Ni, Jinjie and Zheng, Zangwei and Zhou, Wangchunshu and You, Yang},
  journal={arXiv preprint arXiv:2402.01739},
  year={2024}
}

@article{allenai:arc,
      author    = {Peter Clark  and Isaac Cowhey and Oren Etzioni and Tushar Khot and
                    Ashish Sabharwal and Carissa Schoenick and Oyvind Tafjord},
      title     = {Think you have Solved Question Answering? Try ARC, the AI2 Reasoning Challenge},
      journal   = {arXiv:1803.05457v1},
      year      = {2018},
}
